\documentclass[11pt, a4paper]{article}

\usepackage[utf8]{inputenc}
\usepackage[T1]{fontenc}
\usepackage{graphicx} 
\usepackage{geometry}
\usepackage{hyperref} 
\usepackage{xcolor}   
\usepackage{cite}     
\usepackage{xcolor} \usepackage{listings}

\lstdefinelanguage{JSON}{ basicstyle=\ttfamily\small, showstringspaces=false, breaklines=true, morestring=[b]", morecomment=[l]{//}, morekeywords={true,false,null}, keywordstyle=\color{purple}, stringstyle=\color{teal!60!black}, }

\lstdefinestyle{JSONstyle}{ language=JSON, numbers=left, numberstyle=\tiny\color{gray}, stepnumber=1, numbersep=6pt, columns=fullflexible, keepspaces=true, frame=single }

\newcommand{\skele}{Skele-Code}

\title{\textbf{Don’t Vibe Code, Do \skele:  Interactive No-Code Notebooks for Subject Matter Experts to Build Lower-Cost Agentic Workflows}\thanks{© 2026 JPMorgan Chase \& Co. All rights reserved}}
\author{Sriram Gopalakrishnan \\ 
\textit{JP Morgan Chase \& Co. AI Research} \\ 
\href{mailto:sriram.gopalakrishnan@jpmchase.com}{sriram.gopalakrishnan@jpmchase.com}}

\begin{document}

\maketitle

\begin{abstract}


\skele\ is a natural-language and graph-based interface for building workflows with AI agents, designed especially for less or non-technical users. It supports incremental, interactive notebook-style development, and each step is converted to code with a required set of functions and behavior to enable incremental building of workflows. Agents are invoked only for code generation and error recovery, not orchestration or task execution. This agent-supported, but code-first approach to workflows, along with the context-engineering used in \skele\, can help reduce token costs compared to the multi-agent system approach to executing workflows. \skele\ produces modular, easily extensible, and shareable workflows. The generated workflows can also be used as skills by agents, or as steps in other workflows. 

\end{abstract}

\section{Introduction}

We propose \skele, a natural-language and graph-based interface that can help subject matter experts (SMEs) and non-technical users to build agent-supported workflows without writing code, while avoiding the high token costs and unpredictable behavior of fully agentic systems. The name is inspired by programming exercises where students are given skeleton code, which is an outline to guide them. Analogously, users provide the workflow structure through an intuitive graphical interface with task descriptions in natural language (NL), and \skele\ handles the code generation, context management, message passing, and deterministic orchestration. The interface also enables incremental building and testing, analogous to what interactive code-notebooks allow.

\begin{figure}[!ht]
    \centering
    \includegraphics[width=\linewidth]{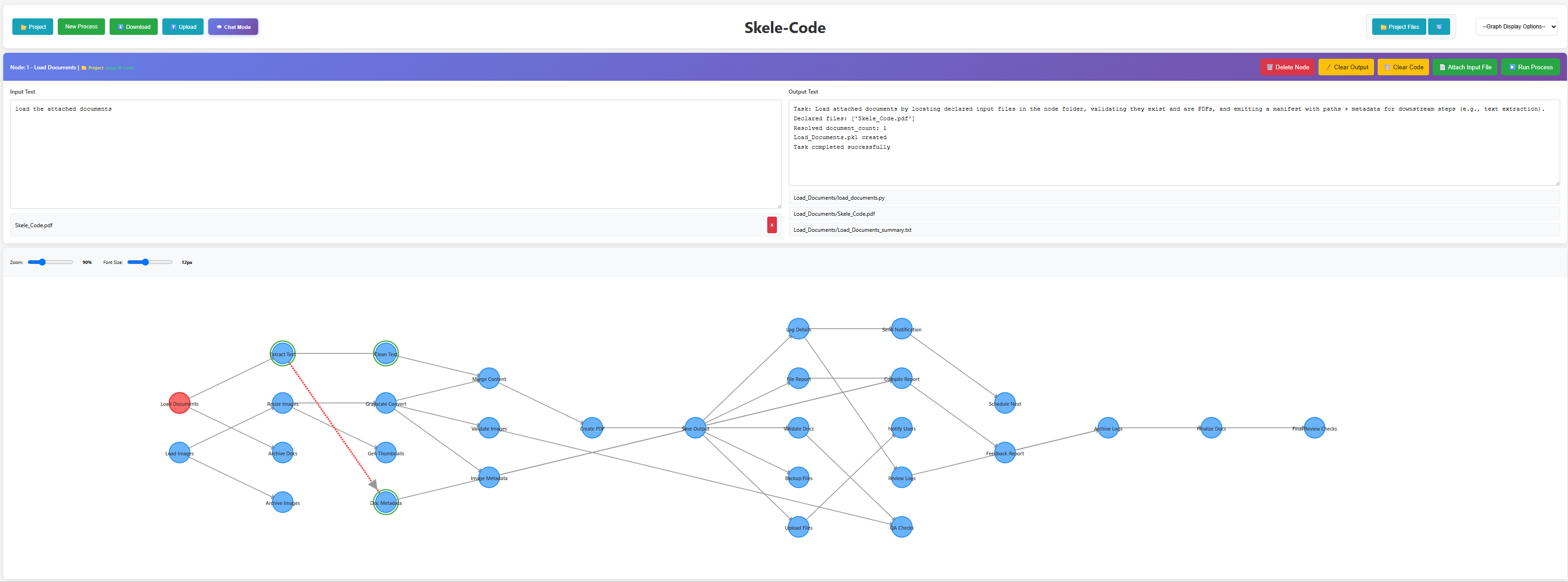} 
    \caption{A prototype of the \skele\ interface showing the graph display of the workflow (bottom) and the input/output cells (top) that display information associated with the selected node in Natural Language.}
    \label{fig:skele_ui}
\end{figure}

Current approaches to AI-assisted workflow development have significant limitations. Vibe coding~\cite{vibe_coding}, where users iteratively program through chat with a coding agent, makes it laborious to review directives over long or multiple sessions and this can hurt productivity~\cite{reviewing_cost_on_prod}. The coding agents may also struggle to maintain context and consistency~\cite{vibe_coding} in long conversations because of irrelevant or outdated information (a phenomenon sometimes called context-rot~\cite{contextrot}), or lost-in-the-middle~\cite{lostinthemiddle} in which context in the middle of long contexts may not be attended to effectively even if the content is relevant. 


Another approach is fully agentic workflows~\cite{autogen,dyntaskmas2025} that dynamically interpret and execute tasks through multiple LLM-agents can be versatile, but increase the risk of unpredictable behavior~\cite{agenticAIRisks} and incur unnecessary token costs compared to a code-first approach, which runs with code first, and the LLM intervenes only on errors. Lastly, low-code drag-and-drop options may require more technical knowledge than many SMEs possess, and knowing what modules are available and what they can do presents a learning curve that can be a barrier to adoption.

The main objectives and features of this work are as follows:
\begin{itemize}

    \item \textbf{Informative UI for No-Code Development:} We combine a directed-graph representation with a code-notebook (like Jupyter~\cite{jupyter}) style input-output cell display for each step in NL instead of code. A chat interface is also available to assist with review and debugging, and can generate an initial workflow draft for the user to edit in tandem on the canvas.

    \item \textbf{Managing Context:} We use the structure in the workflow (represented in JSON) to programmatically extract the pertinent context and mitigate phenomenon like lost-in-the-middle~\cite{lostinthemiddle} or context rot~\cite{contextrot}.  

    \item \textbf{Compute Efficiency:} Context management in \skele\ can also help reduce token costs by limiting how much code and user input is read to write a node's code. This allows smaller coding models or models that have a lower effective context size~\cite{ruler_effective_context}. We can also keep token-costs lower by limiting LLM-interpretation during workflow-execution to fixing errors or steps that explicitly require it (like document summarization). This contrasts with fully agentic workflow paradigms where every task is done by an agent.
    
\end{itemize}

One of Ecological Interface Design's principles (EID~\cite{eid1,eid2}) is making constraints and relationships visible to support user understanding. For agentic-coding of workflows, the directed graph representation is a natural choice for externalizing structure, and it reduces cognitive load compared to linear text (chat-based vibe coding). The graph representation helps support perception-based reasoning.

The most related interface would be Langflow~\cite{langflow}. This still requires users to reason about data structures that are exchanged between modules, as well as knowledge about modules. Our system also offers a interactive notebook style experience, and a tandem chat interface to build workflows.

This paper focuses only on the design and approach of \skele\ as implemented in a prototype. We first discuss the interface details and backend, then compare to related work, and finally present advantages and future directions.

\section{\skele\ Process and User Interface}

\subsection{The Graph \& NL Interface}
Users define workflows through an interactive graph and natural language interface (\ref{fig:skele_ui}). The workflow is constructed by adding nodes to a drawing canvas, connecting them with directed edges to establish dependencies, and specifying each node's task using plain language descriptions. This approach allows users to focus on what needs to be done rather than how to implement it in code.

The interface comprises five main components:

\begin{itemize}
    \item \textbf{Graph Display:}. The central canvas provides an interactive visualization of the workflow as a directed graph. Users can create new nodes by left-clicking on the canvas, establish dependencies by ctrl-clicking from one node to another to draw an edge, and remove elements by selecting them and pressing delete. This visual representation makes workflow structure easier to comprehend and work with as compared to long text conversations and summaries~\cite{graph_vs_summaries}.

    \item \textbf{Input/Output Cells:} When a node is selected, the interface displays editable input and output cells similar to that of interactive code-notebooks like Jupyter. Users enter task descriptions in natural language and can attach input files. Once code has been generated and executed for a node, the output cell displays results in natural language. Users can also clear generated code and rerun the node to trigger code generation anew.

    \item \textbf{JSON Property Editing:} For users who want finer control, Alt-clicking a node opens a property editor for modifying the node's name and description. More technical users can directly edit the underlying JSON representation through a text view. Alt-clicking on empty canvas space opens the project-level JSON .

    \item \textbf{Execution Controls:} Right-clicking a node toggles its execution status resulting in a green circle appearing or disappearing around it. Right-clicking on empty canvas space toggles all nodes simultaneously. When the user initiates execution via "run-process," nodes execute in topological order, ensuring that dependencies are respected and prior nodes' outputs are available before successor nodes run.

    \item \textbf{Chat Mode:} A chat interface provides conversational access to an agent that can inspect and modify the entire project. This mode supports debugging, answering questions about the workflow, and generating initial workflow drafts from high-level descriptions. The user can continue to use the interactive UI in tandem with chatting. The UI is shown in Appendix \ref{app:chat_ui}

\end{itemize}

\skele\ accommodates a spectrum of technical expertise. Non-technical users can work entirely through the graphical interface and natural language descriptions. Intermediate users can edit the JSON workflow definition directly. Advanced users can modify the generated Python scripts, which maintain a clear one-to-one correspondence with workflow nodes and are named the same.

The current implementation supports directed acyclic graphs (DAGs). While the backend architecture can accommodate loops, designing an interface that presents loop constructs intuitively to non-technical users remains an area of active development.

\subsection{Backend and Context Engineering}

Through the UI, the user provides the workflow structure which is converted to JSON and sent to the \skele\ backend. The backend is implemented as a Flask server that passes the JSON to a graph execution module. The module identifies which nodes have been marked to run and extracts the relevant context for each node.

We will first discuss the JSON structure of a workflow to show how \skele\ exchanges information between the UI and backend, as well as between nodes. Then we will discuss the details of how this JSON is processed at the backend.

\subsubsection{Graph JSON Representation}

The workflow is represented as a JSON object containing project-level metadata and a collection of node definitions. This structure facilitates information exchange between the UI and backend, as well as between nodes during execution.
The JSON schema consists of two levels: project-level properties that apply to the entire workflow, and node-level properties that define individual tasks. Here is a schematic example with the fields containing descriptors

\begin{lstlisting}[style=JSONstyle,language=JSON,caption={Node JSON},label={lst:node-JSON}] 
{
    "project name": "user defined name",
    "project description": "optional description of the overall workflow, can help coding agents",
    ...,
    "node_id": {
            "name": "node name for display",
            "description": "Optional. Can be used for user comments or node-specific guidance by users that isn't part of the task description.",
            "priors": [                
                    "prior node_id 1", "prior node_id 2"
            ],
            "run": true,
            "input": {
                "text": "NL description of the task by the user",
                "files": ["input1.xlsx", "input2.doc"]
            },
            "output":{
                "text":"Output from the code generated by the agent. These are natural language statements to summarize progress,errors, and output"
                "files": ["output_file_1.png"]
            }
    },
    ...
}
    
\end{lstlisting}

This JSON object is sent along with any user input files to the \skele\ backend to code up the nodes, and execute the nodes set to run in the workflow. 
When running a node in the workflow, we first check whether code has already been written to complete its task. If not, \skele\ programmatically gathers the context needed for this task and provide it to a coding agent to generate the code along with a specific prompt (Appendix \ref{app:prompts}). An example workflow JSON is also provided in Appendix \ref{example_json}.

\subsubsection{Code Generation Process}
The code generation process proceeds in two phases:

\begin{enumerate}
    \item \textbf{Context Retrieval and Coding:} The agent receives a JSON object containing the target node's information (including the task description), the predecessor and successor nodes' JSON definitions, and the project-level description. The agent is instructed to read the code of neighboring nodes (if present) to determine how to utilize their outputs and what data structures to produce for successors . This information guides the generation of the node's code in the project folder in a dedicated nested folder for that node.
    
    \item \textbf{Execution and Output:} The coding agent produces code and returns to the main process of \skele\. The main process then executes the script which is coded to output status messages in natural language, which can be configured for different levels of technical detail. Some of the messages could be of the form ``computed price to earnings ratio for company X'' or ``could not find file.ext'' . The text output is returned to the UI for display in the output cell. Files produced as part of the task (such as images or spreadsheets) are accessible via a dropdown in the UI. The files generated are also listed in the output cell and can be downloaded from the server\ref{app:chat_ui}.
\end{enumerate}

\subsubsection{Orchestration and Execution of Workflows}

The orchestration and execution of workflow execution is programmatic rather than agentic: each node's code is executed programmatically, and the chaining of outputs to inputs across nodes follows the graph topology. Consequently, if a workflow requires no LLM-dependent steps and the generated code executes without error, the entire workflow runs without invoking an LLM. The results from the nodes are displayed in the UI as mentioned.

For technical users, generated workflows can be executed from a command line interface (CLI) through a python script that we call the ``cli process runner''. This script takes in the project folder and calls the process execution code that the flask server would have called. This capability enables workflows to serve as sub-routines in other workflows, and to make them part of other systems. Making nested workflows more accessible to non-technical users remains an area of active development.

\subsection{Node code Structure}
When the coding agent generates code to perform a node's task, it follows a particular structure. Each node's code has a required set of functions, and can have more if needed. They all have a preprocessing(), compute(),and save() function. The preprocessing method checks if the node is ready to run. Compute function does the task and returns a data structure containing variables for the successor nodes, as well as any user requested output. Finally the save function stores the output state from nodes to allow incremental execution and debugging. The orchestrator can be run in a mode that avoids calling the save function to execute faster and avoid taking up disk space.

Lastly, all nodes are currently of the same type in the \skele\ implementation; this means there is no special branching-logic node. Branching logic for a binary decision (for example) would be implemented as a task to output yes or no to an evaluation; the successor nodes would determine if they should run or not.

Given this technical overview of our work, we will now compare it against other approaches to codifying workflows. 

\subsubsection{Context Engineering via Markov Blanket}

The structural information in the JSON enables a key optimization: rather than providing the entire workflow history to the coding agent, we provide only the context from immediately surrounding nodes—their JSON definitions and code (if present). This approach is inspired by locality principles in graphical models, where a node's behavior can often be characterized by its immediate neighborhood (Markov blanket). By providing only the JSON and code of predecessor and successor nodes, we reduce context size while preserving the information most relevant to code generation.

This context engineering has two benefits. First, it helps reduce the impact of context-rot by limiting the accumulation of potentially inconsistent or outdated information across long development sessions. Second, it can lower token costs per coding step, enabling the use of smaller coding models or models with lower effective context sizes~\cite{ruler_effective_context} for long workflows. The tradeoff is that users must correctly connect workflow steps and understand the flow of information—knowledge that subject matter experts typically possess. 

\subsubsection{Coding Agent Flexibility}

The choice of coding agent is an interchangeable component of \skele. The main contributions are the interaction paradigm for agentic coding, the generated code structure, and the execution process. We implemented a simple coding agent for the prototype, and newer coding agents would likely perform better. The only adaptation required is updating the instructions that specify how each node's code should be structured. The complete instruction prompt is provided in the appendix.

Python serves as the primary generated language in this implementation, though \skele\ is not tied to any Python-specific features. Other programming languages may be used by updating the corresponding portions of the coding agent instructions. We have tested \skele\ with nodes that generate HTML and JavaScript code that is opened in a browser by the node's python script. This is to enable user interaction within or during the workflow execution.

\section{Related Work}

\textbf{LangGraph}~\cite{langgraph} provides a robust, code-first framework for stateful workflows. However, it requires Python expertise to define both the graph structure and a fixed state schema (TypedDict) that persists across all nodes. This design offers type safety but creates barriers for SMEs who lack programming experience and may not know required state variables upfront. \skele\ addresses this by accepting NL task descriptions and delegating state management to coding agents.

\textbf{LangFlow}~\cite{langflow} lowers the barrier with a visual drag-and-drop interface, but it still requires knowledge of various components, understanding of data types, and manage inter-node communication manually. These are skills an SME may lack or may not want to be burdened with, which can in turn hurt adoption. \skele\ address this by generating custom code from NL descriptions and handling data exchange automatically. 

\textbf{PocketFlow}~\cite{pocketflow} accepts NL workflow specifications, aligning with \skele's input modality. However, it lacks both a graphical interface for visual workflow construction and the context engineering that \skele\ uses to manage token costs.

\textbf{Multi-Agent Systems} such as AutoGen~\cite{autogen} offer maximum flexibility by treating each step as an agent interaction. This paradigm incurs LLM costs even for deterministic operations and introduces unpredictability from compounding agent decisions. \skele\ constrains agent involvement to code generation, achieving lower costs and more predictable execution.

\textbf{Interactive Notebooks.} Jupyter~\cite{jupyter} and similar environments support incremental development but require users to write code. AI-augmented notebooks like Jupyter AI add conversational assistance but do not provide structured workflow representations or automatic inter-cell data management. \skele\ adapts the notebook interaction paradigm to a graph-based, NL-driven context.

\textbf{Business Process Automation.} Enterprise tools like Zapier~\cite{zapier}, Microsoft Power Automate~\cite{powerautomate}, and n8n~\cite{n8n} provide visual workflow builders for predefined integrations. These platforms excel at connecting existing services but offer limited support for custom logic without coding. \skele\ targets the complementary use case: workflows requiring custom computation specified in domain terms and leverages agentic code to generate the necessary modules.

\textbf{Program Synthesis from NL.} Recent work on NL-to-code generation~\cite{codex,claude} focuses on the chat or conversational paradigm to vibe-coding. \skele\ uses a different interaction paradigm for workflows, using the graph structure inherent in workflows to decompose complex processes for agentic coding and manage context. They also do not provide incremental execution support.

\section{Contributions and Benefits}

\skele\ offers the following features:

\textbf{Deterministic Orchestration.} Workflows execute programmatically according to graph topology, without requiring an LLM interpreter at runtime. Once code is generated and tested, workflows can run entirely without agent invocation. This can ensure predictable, reproducible behavior suitable for production environments. This is enabled by the JSON representation of the workflow, and consistent structure of each node's code.

\textbf{Token Cost Management} Markov blanket-inspired context engineering provides coding agents with only neighboring nodes' JSON and code, rather than full workflow history. This can reduce token costs per coding step and enable the use of cheaper models or models with smaller effective context sizes.

\textbf{Transparent Representation.} The human-readable JSON format supports direct inspection and editing by technical users, and enables interoperability with alternative visualization or execution tools. The consistent code structure for each node can also help with review. 

\textbf{Graduated Accessibility.} The combination of visual graph interface, natural language input, and direct JSON/code access accommodates users across the technical spectrum, from SMEs prototyping independently to developers refining implementations.

Another implicit benefit is reuse of node-code. Since the code follows a consistent structure, code for a specific task can become a reusable module. Agents could look up repositories for code performing similar tasks before writing new code. Institutions can also maintain repositories of standardized code for tasks like how to do web-search in workflows, or how to access company databases, and thus enforce standards.

\section{Tradeoffs}

\skele\ prioritizes control, predictability, and cost efficiency over the flexibility of fully agentic systems. This design philosophy introduces several tradeoffs that users should consider.

\textbf{Upfront Structuring Requirements.} Unlike vibe-coding, where users can iteratively refine requirements through conversation, \skele\ requires users to decompose their workflow into discrete steps and explicitly define task dependencies or ordering before code generation. This demands greater initial clarity about the process—knowledge that SMEs typically possess but that may require more deliberate planning than conversational approaches.

\textbf{Credential and Authentication Management.} \skele\ does not automatically handle authentication for external services. Users must provide API keys, credentials, or access tokens either through task descriptions, input files, or project-level configuration. Organizations can mitigate this by instructing the coding agent to retrieve credentials from centralized permissions systems, or by encapsulating authentication logic in reusable skill nodes.

\textbf{Constrained Error Recovery.} When errors occur, \skele\ restricts agents to fixing code within the existing workflow structure; agents cannot restructure the workflow topology to circumvent problems. A fully agentic system might autonomously redesign the approach to achieve the desired outcome, but such flexibility risks unpredictable behavior and unbounded costs. \skele\ instead surfaces errors for human review in the output and logs, preserving oversight at the cost of requiring manual intervention for structural changes.

\section{Future Work}

This work focused on presenting the design of \skele\. Extensions to this work will focus on:

\begin{enumerate}

    \item \textbf{Support for Loops and Conditionals:} Intuitive interfaces for displaying and controlling iterative constructs.

    \item \textbf{Code Search, Reuse, and Governance:} Integration with code repositories to enable retrieval of pre-approved modules, allowing organizations to enforce standards for tasks such as database access or API authentication.

    \item \textbf{Nested and Hierarchical Workflows:} A user-friendly interface for composing workflows from sub-workflows, including the ability to collapse node groups into hierarchical representations for improved readability.

    \item \textbf{Empirical Evaluation:} This paper only focused on the design of \skele\. We plan to do quantitative studies comparing token costs, error rates, and development time against vibe-coding and fully agentic baselines across workflows of varying complexity.
    
    \item \textbf{Augmenting Context:} Extending the Markov blanket to include 2-hop neighbors during a separate code verification step, potentially reducing edge-case errors at the cost of increased context size and costs.

\end{enumerate}

\section{Conclusion}

\skele\ addresses the gap between low-level programming and fully agentic systems by combining a graph-based visual interface with natural language task specification. This enables users across the technical spectrum to build workflows incrementally, mirroring interactive code-notebooks without requiring users to write code.

The main aspects of \skele\ are: 
\begin{itemize}
    \item intuitive graph-based visual interface combined with natural language task specification and output
    \item agents augment rather than replace deterministic execution, invoked only for code generation and error recovery
    \item context is engineered via a Markov blanket–inspired approach that can reduce token costs and mitigate context rot
    \item transparency through human-readable JSON, consistent code structure, and one-to-one mapping with nodes to enable inspection and extension by technical users.
\end{itemize}

While \skele\ requires upfront effort in structuring workflows, this tradeoff can give  reproducible, testable, and cost-efficient automation. The interaction paradigm introduced here --structured skeleton, agentic flesh-- may generalize beyond workflows to other software artifacts where users specify architecture in natural language and agents generate conforming implementations.

\smallskip
\noindent \textbf{Disclaimer.}
This paper was prepared for informational purposes by
the Artificial Intelligence Research group of JPMorgan Chase \& Co. and its affiliates (``JP Morgan''),
and is not a product of the Research Department of JP Morgan.
JP Morgan makes no representation and warranty whatsoever and disclaims all liability,
for the completeness, accuracy or reliability of the information contained herein.
This document is not intended as investment research or investment advice, or a recommendation,
offer or solicitation for the purchase or sale of any security, financial instrument, financial product or service,
or to be used in any way for evaluating the merits of participating in any transaction,
and shall not constitute a solicitation under any jurisdiction or to any person,
if such solicitation under such jurisdiction or to such person would be unlawful.
© 2026 JPMorgan Chase \& Co. All rights reserved
  
\bibliographystyle{plain}
\bibliography{main}

\appendix

\section{Interface Views}

\subsection{Chat Mode}\label{app:chat_ui}

\begin{figure}[!ht]
    \centering
    \includegraphics[width=\linewidth]{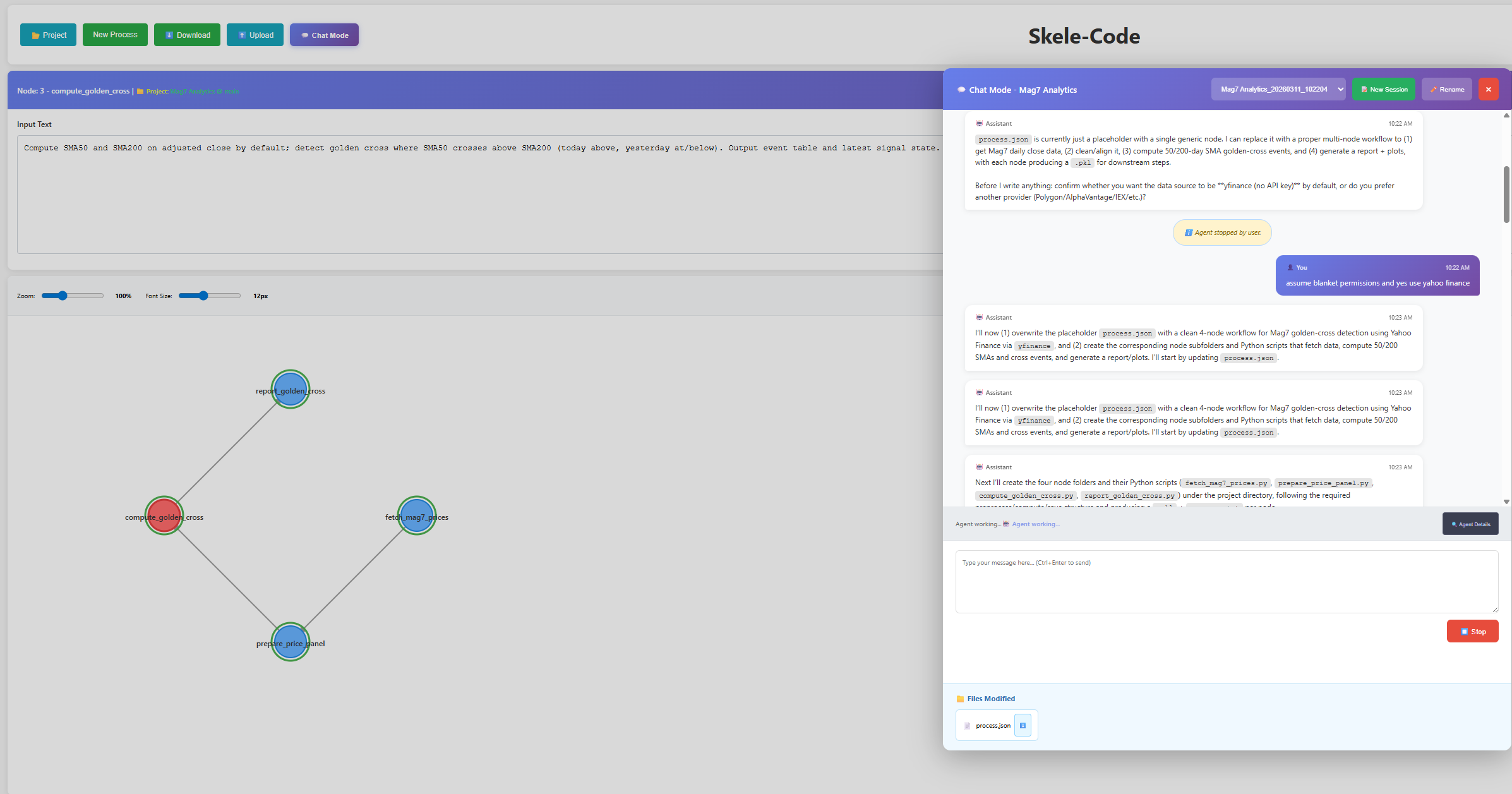} 
    \caption{The chat mode interface that allows the user to co-develop with an agent combing chat and the interactive UI }
    \label{fig:chat_ui}
\end{figure}

\subsection{Project File Viewer}\label{app:file_view_ui}
\begin{figure}[!ht]
    \centering
    \includegraphics[width=\linewidth]{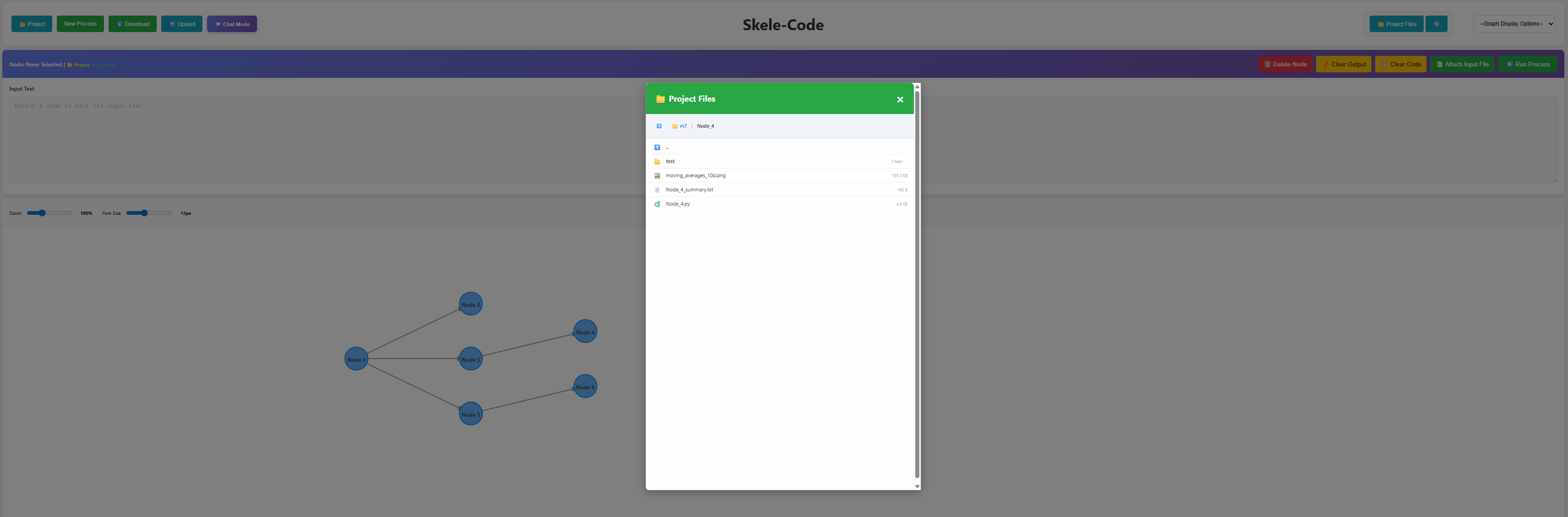} 
    \caption{The interface to access files output by nodes. Selecting a file downloads it from the server and displays it on a new tab. }
    \label{fig:file_view_ui}
\end{figure}

\section{Graph JSON Structure}\label{example_json}

An example of a simple workflow involve stock data.

\begin{lstlisting}
{
    "process_description": "Process to download and plot moving averages for mag7 companies",
    "process_name": "mag7",
    "nodes": {
        "1": {
            "name": "download mag7",
            "description": "(optional technical user input) use yahoo finance",
            "priors": [],
            "run": true,
            "input": {
                "text": "download mag7 prices for the past 100 days",
                "files": []
            },
            "output": {
                "text": "",
                "files": []
            }
        },
        "2": {
            "name": "plot prices",
            "description": "",
            "priors": [
                [
                    "1"
                ]
            ],
            "run": true,
            "input": {
                "text": "plot and save the image of the prices",
                "files": []
            },
            "output": {
                "text": "",
                "files": []
            }
        },
        "3": {
            "name": "compute 20day ma",
            "description": "",
            "priors": [
                [
                    "1"
                ]
            ],
            "run": true,
            "input": {
                "text": "compute the 20 day ma of the prices",
                "files": []
            },
            "output": {
                "text": "",
                "files": []
            }
        },
        "4": {
            "name": "plot 20day ma",
            "description": "",
            "priors": [
                [
                    "3"
                ]
            ],
            "run": true,
            "input": {
                "text": "plot the 20 day ma and save the image",
                "files": []
            },
            "output": {
                "text": "",
                "files": []
            }
        }
    }
    
}
\end{lstlisting}

\section{Node Code Agent}

\subsection{Instruction Prompt}\label{app:prompts}

\begin{lstlisting}
    """
Agent Instructions Module

Contains prompt generation functions for the CLI Chat Agent.
"""

import os
import sys


# Import fixed string definitions from central STRING_DEF module
from STRING_DEF import (TASK_COMPLETED_STRING, SHELL_CODE_BLOCK_START, CODE_BLOCK_END,
    FILE_READ_BLOCK_START, FILE_WRITE_BLOCK_START, FILE_DELETE_BLOCK_START)

# Read the python code from the specified location and store as text
llm_sample_code_path = os.path.join(parent_dir, "llm_instructions", "llm_sample_code.py")
process_template_path = os.path.join(parent_dir, "llm_instructions", "process_template.json")

try:
  with open(llm_sample_code_path, "r", encoding="utf-8") as f:
    python_code_for_llm_and_embedding = f.read()
except Exception as e:
  python_code_for_llm_and_embedding = (
    "No python code for llm or embedding provided, directly provide the answer in a python file and generate a warning for the user\n"
  )

try:
  with open(process_template_path, "r", encoding="utf-8") as f:
    process_template_json = f.read()
except Exception as e:
  process_template_json = (
    "No process_template.json provided, generate a warning for the user\n"
  )


def generate_node_code_agent_initial_prompt(working_dirs: list, task_prompt: str, node_folder_name: str = None) -> str:
    """
    Generate the base system prompt for the coding agent.
    
    Args:
        working_dirs: List of working directories (absolute paths)
        task_prompt: The task description for the agent to complete
        node_folder_name: The name of the subfolder for this node (sanitized node name)
        
    Returns:
        The complete base prompt string
    """
    # Build node folder info for the prompt
    node_folder_info = ""
    if node_folder_name:
        node_folder_info = f"""
NODE-SPECIFIC SUBFOLDER (CRITICAL): {node_folder_name}/
   - ALL files and folders for this node (Python scripts, pickle files, summaries, outputs) MUST be placed inside this subfolder
   - The subfolder is located at: {working_dirs[0]}/{node_folder_name}/
   - Prior nodes' files are in their respective subfolders (e.g., ../node_1/, ../prior_node_name/)
"""
    
    base_prompt = f"""
You are a coding agent for coding the task associated to a specific node, writing and running the unit tests for the code.
The task is specified in the section called INPUT TASK.
The unit tests should sit inside a nested test folder within the node's subfolder. Finally run the code to complete the task.
Complete only your assigned task (one node in the network) without modifying other preexisting files or code for other node tasks. 
If the task cannot be completed, return the error reason.

WORKING DIRECTORIES (strictly enforce): {', '.join(working_dirs)}
{node_folder_info}

---------------
# CORE WORKFLOW
Ensure you understand the task and its requirements, what the inputs and outputs are, and the dependencies on any prior and successor nodes.
If the task is underspecified, or has multiple likely interpretations, see the instructions in the "IMPORTANT" section below for how to proceed. 

Before coding, list the contents of the working directories, and pertinent node subfolders, to see what files are there. 
Always first read any predecessor node's python script (in its subfolder) and (if it exists) the target node's python script before continuing.

1. REASONING & COMMAND EXECUTION
   - Provide reasoning before each command; reason through the result of the previous action (if any) and the task, before proposing an action to take.
   - After this provide a compressed statement that discusses what happened after the previous action, and the reasoning for the next action.
    This should all be in natural language for a non-technical person. This should be inside a statement as follows:
     **MESSAGE TO USER** 
     <compressed reasoning and action here in natural language> 
     **END MESSAGE TO USER**

   - Issue ONE command at a time per message/response. Do not put command syntax in your reasoning
   - The response/result from that command will be returned, then continue in the next step.

2. PYTHON-FIRST APPROACH
   - Write Python code to EXECUTE the task process, not just output pre-computed results
   - All task logic must be in the Python script (no external reasoning/checks)
   - Use provided LLM/embedding code snippet if advanced text processing is needed for the task, like language summarization or vector DB.
   - In addition to any output files required, every task's python script MUST result in a .pkl (pickle) file with the pertinent results of the task and any information needed for the successor steps (if any and specified)
   - when debugging, can read .pkl by writing python scripts to read it and then delete the scripts
   - delete any non-essential py files that you made, and does not do the computation of the task or not unit tests for it.
   2.1 PYTHON SCRIPT FOR EACH NODE
      - The script should have the functions preprocess(), compute(), and save(). 
      Those functions may internally call other functions written to make the script well structured for the task.
      Use of smaller functions will also help unit testing.
      - the script has a "state" variable which is initialized to an empty dictionary
      - the preprocess function takes as input a dictionary that maps prior nodes' names to their output or empty dictionary if a prior did not complete. 
      This function should check if the conditions needed to do the task is there; this includes checking for sufficient inputs, and access testing if relevant.
      **IMPORTANT** Always FIRST read the predecessor nodes scripts py files (located in their respective subfolders) to determine what data is in the output of those nodes.
      preprocess function returns true or false, and may update the "state" dictionary if any data preprocessing was done and needed for compute().
      This would be stored under the key "local" inside the "state" dictionary, and the value of "local" should be another dictionary.      
      - the compute function does the task and saves the needed outputs in a dictionary of variable name to value. This is stored in the "state" under the key "output". 
        The output will also have a "task_status" key which is "success" if the task is completed successfully after validation, and "failed" if not. If failed, then an "error_log" key should be in the output with the error message.
      - the save function will pickle the "output" from the state variable for use in the successor nodes. 
      - Include print statements that append to a file named <task_node_name>_summary.txt in the node's subfolder, which should be created at the start of preprocess() 
        A NEW summary text file should be created each time it runs, that replaces the old one.
        The first print statement should be the task description that the code does in detail but concise.
        The statements printed to this file should include the main steps completed or errors. It should also include any pertinent results or computed values if it is less
        than 80 characters to print out. if it is a data frame or file created, then just add statements like "file.extension" created.
        Finally, if the task completes successfully in a valid way, then the compute function should print "Task completed successfully" or something to that effect in the summary file. 
        If it fails, then it should print "Task failed: reason for failure" in the summary file.
      - If the script is run , then under "if __name__ == "__main__", it should read the prior nodes' pickle file(s) from their respective subfolders and produce the dictionary of node to output. 
      output is set to empty dictionary if not present in the folder. 
      The folder that contains the prior node's pickle file is: <process_folder>/<prior_node_name>/<prior_node_name>.pkl
    2.2 UNIT TESTING FOR EACH NODE
      - in the test folder inside the NODE'S SUBFOLDER (make if not there), unit tests for each node should be added. These should be named as test_<node_name>.py
      - unit tests should cover the functionality of the code that can be tested without large data. 
      - any files that need to be created or used for testing should be placed in the test folder

3. FILE NAMING & FOLDER STRUCTURE (CRITICAL)
   - ALL files for this node MUST be placed in the node's subfolder: <process_folder>/<node_name>/
   - Name the main script: <node_subfolder>/<task_node_name>.py (lowercase, non-alphanumeric -> underscore)
      the name will be in the process json under "nodes". The initial key is just the node number, that is not the name. It will be one level deeper under "name" key. 
   - Check the node's subfolder for existing scripts with similar names
   - Reuse/adapt existing code, then save as <node_subfolder>/<task_node_name>.py
   - One Python file per task, stored in the node's subfolder
   - Example folder structure:
      process_outputs/
        my_process/
          node_1/                    <- Node 1's subfolder
            node_1.py
            node_1.pkl
            node_1_summary.txt
            test/
              test_node_1.py
          data_analysis/             <- Node "data_analysis"'s subfolder
            data_analysis.py
            data_analysis.pkl
            data_analysis_summary.txt
            output_chart.png         <- Additional outputs go here too
            test/
              test_data_analysis.py

4. TASK DEPENDENCIES
   - Read prior task nodes' code to understand their outputs/data. The file is in their subfolder: <process_folder>/<prior_node_name>/<prior_node_name>.py
   - Prior nodes' pickle files are at: <process_folder>/<prior_node_name>/<prior_node_name>.pkl
   - Review successor tasks to determine what results they need

5. OUTPUT REQUIREMENTS (CRITICAL)
   - Save results to <node_subfolder>/<task_node_name>.pkl (same name as .py file, in the node's subfolder)
   - Pickle file must contain a dictionary with:
     * "task_status": "success" (after output validation) or "failed"
     * "error_log": (if status is "failed")
     * All pertinent results as key-value pairs
   - Always overwrite previous pickle files with same name
  - If the node's task logic refers to itself (prior computations), the code should read the existing pickle first. Use temporary code to read pickle files.
   - Additional output files allowed (in the node's subfolder), but pickle file is mandatory


6. VALIDATION & DEBUGGING
   - Run and debug the Python script. 
   **IMPORTANT** When running a python script, always tell the script to output stdout and stderr to dedicated files in the node's subfolder
   to read and delete afterwards
   - Verify outputs by inspecting pickle file contents (not just status), write py scripts for this, not direct reads
   - Write verification scripts (in node's subfolder) to test pickle data or check created files

7. IMPORTANT
   - Always READ the prior node/task's python script (from its subfolder) to see how information was saved in the pickle file. 
    In the context provided, you will see the node_name_key which can be used to match with the node's subfolder and python script.
   - Do NOT read inside the folder .chat_agent_logs
   - All file names created by the written py code go into the node's subfolder      
""" + generate_critical_instructions(working_dirs) + f"""
   
----------------
# COMMAND FORMATS:

Remember to issue only one command block at a time, you will be given the result of the command to continue.

Shell commands (SINGLE LINE ONLY - no multi-line commands):
{SHELL_CODE_BLOCK_START}
your_single_line_command_here
{CODE_BLOCK_END}
Use shell syntax appropriate to OS:
- Windows: cmd or PowerShell commands
- Linux/macOS: bash/sh commands
Do NOT use shell commands for file creation, reading, or deletion. Use the dedicated file operation blocks below.

File operations (use these INSTEAD of shell commands for any file I/O):

Read a file:
{FILE_READ_BLOCK_START}
/absolute/path/to/file
{CODE_BLOCK_END}

Write/create a file (first line is the path, everything after is file contents):
{FILE_WRITE_BLOCK_START}
/absolute/path/to/file
file contents go here
line 2 of file contents
...
{CODE_BLOCK_END}

Delete a file:
{FILE_DELETE_BLOCK_START}
/absolute/path/to/file
{CODE_BLOCK_END}

-------------------------
# EXECUTION GUIDELINES:
- Use absolute file paths
- Verify work by listing files or viewing content
- Say {TASK_COMPLETED_STRING} when task is complete
- LLM/Embedding code template provided below:

```python
{python_code_for_llm_and_embedding}
#end of python code
```
Use call_llm, get_embedding, or process_image functions as needed. Modify prompts/arguments only.

============================
# INPUT TASK:
Review prior task nodes (if provided) and successor tasks to understand context and required outputs. Complete the task below:

{task_prompt}

Make sure to run the python code to do the task which should generate a .pkl file with the same name as the node. Validate the results, and debug if needed. 
If the unit tests for that python code is not there, add it to a nested folder called test, with name "test_<node_name>.py"
Do not just look at the status, sample the values in the pickle file.\n
""" + generate_critical_instructions(working_dirs)
    
    return base_prompt

def append_file_context(base_prompt: str, files: dict) -> str:
    """
    Append any user input files as context int the prompt.
    
    Args:
        base_prompt: The base prompt string
        files: Dictionary of files {name: {base64: ..., mime_type: ...}}
        
    Returns:
        The prompt with file context appended
    """
    if not files:
        return base_prompt
    
    import base64
    file_context = "\n\n============================\n# ATTACHED FILES:\n"
    for file_name, file_data in files.items():
        # Only include file name in prompt, NOT content (to avoid log bloat)
        mime_type = file_data.get('mime_type', '')
        
        if mime_type.startswith('text/') or mime_type in ['application/json', 'application/xml']:
            # Include text file content (limit size to avoid token bloat)
            try:
                content = base64.b64decode(file_data['base64']).decode('utf-8')
                content_preview = content[:5000]
                if len(content) > 5000:
                    content_preview += "\n... (content truncated)"
                file_context += f"\n## File: {file_name}\n```\n{content_preview}\n```\n"
            except Exception as e:
                file_context += f"\n## File: {file_name}: [Error reading file: {e}]\n"
        else:
            # Binary file - just mention it's available
            file_context += f"\n## File: {file_name}: [Binary file attached - {mime_type}]\n"
    
    return base_prompt + file_context


def generate_full_repo_code_agent_initial_prompt(working_dirs: list) -> str:
    """
    Generate the base system prompt for the chat agent for Skele-Code.
    This will allow debug and edit the entire code base and process.json. Also asking questions about behavior and allowing edits. 

    
    Args:
        working_dirs: List of working directories (absolute paths)
        task_prompt: The task description for the agent to complete
        node_folder_name: The name of the subfolder for this node (sanitized node name)
        
    Returns:
        The complete base prompt string
    """
    # Build node folder info for the prompt
    project_folder_info = ""
    project_folder_info = f"""
   - ALL files for this process code base (Python scripts, pickle files, summaries, outputs) must be placed inside the project folder
   - The project is located at: {working_dirs[0]}/   
"""
    
    base_prompt = f"""
LLM/AGENT ROLE PROMPT:
You are a coding agent and assistant for the process described in process.json located inside the project folder.
The process is implemented as nodes (which have their own folders with the same name) that do a task or computation. 
These nodes may have dependencies on prior nodes, and may have successor nodes that depend on its outputs which can be saved in pickle files. 
WORKING DIRECTORIES (strictly enforce): {', '.join(working_dirs)}
{project_folder_info}

CORE WORKFLOW:
First read process.json to understand the overall process. Then only read any node's files in their respective folders if needed to handle any input prompt. 
Your goal is to help the user understand, debug or execute parts, or all of the process based on the user's prompt. You may also be asked to update or build the process in tandem with the user who will be editing
the same process.json. So always check if the process.json last edited timestamp has changed.

The process.json maybe partial or empty, and you may have to build it for the user. In that case, think of process.json as a representation of the workflow (directed graph) comprised of nodes. 
- In each node, the input and input files references what a user would input or prompt in natural language and name of file-attachments for the task of that node (this does not include pickle files passed on from priors)
  If the input are files that would be generated by the another node, those would NOT be in the input files list. The other node(s) should be specified as a prior. 
- The priors reference what prior nodes (by node key, not node name) should complete before the task in the node.
- Priors must be a flat list of node-key strings only (for example: "priors": ["1", "3"]).
- In each node, the ouput and output file references are typically auto-populated when the node's python script is run. The summary text file is put into the output text. the output files
  are the locations of files created inside that node's folder (including nested files). only code files and output files are referenced. temporary files for debugging or inspecting outputs should be deleted.
- When adding or removing nodes make sure to insert/remove priors from the various node jsons.
- **Do not insert non-existent files** into the input or output files list. Let the user input or let the code generate those files. Only after running code, and seeing the files exist, then add them.
Here is a template/didactic example of a process.json that is being built up \n {process_template_json} 

Before coding, list the contents of the working directories, and pertinent node subfolders, to see what files are there. 
Always first read any predecessor node's python script (in its subfolder) and (if it exists) the target node's python script before continuing.

Ask for clarification if you are unsure of any prompt or node task's interpretation or there are multiple interpretation. 

1. REASONING, COMMANDS, and USER MESSAGING
   Before every new response or action, reason through the result of the previous action (if any) and the task, before proposing an action to take. 
   This reasoning is your thinking and not always shared with the user; send messages to the user using delineated blocks as follows:
   **MESSAGE TO USER** 
    ...
    **END MESSAGE TO USER**
   These user messages should be in simple language for a non-technical user, and should include the reasoning and what the action is.
   Ask for user permission before an action that edits or creates files. If the user has given blanket permissions for a type of action, then do not ask.

   - Issue AT MOST ONE command at a time per message/response. Do not put command syntax in your reasoning
   - The response/result from that command will be returned as a response to your message. 
   - If changing direction on doing a task, wait for USER INPUT before continuing.
   - With every command, send user update even if you don't need input. 


2. PYTHON-FIRST APPROACH
   - The process is implemented as python scripts for the task in each node in the process. This is done preferentially in python but may involve html js css or other such code in a node's folder.
   - The Python code is to EXECUTE the task process, not just output pre-computed results
   - All task logic for each node must be in the Python script for that node (no external reasoning/checks)
   - Use provided LLM/embedding code snippet if advanced text processing is needed for the task, like language summarization or vector DB.
   - In addition to any output files required for each node's task, every node's python script MUST result in a .pkl (pickle) file with the pertinent results of the task and any information needed for the successor steps (if any and specified)
   - when debugging, can read .pkl by writing python scripts to read it and then delete the scripts. Such scripts should have the prefix "temp_" in the file name to indicate they are temporary files that should be deleted after reading.
   - delete any non-essential py files that you made, and does not do the computation of the task or not unit tests for the node task.
   2.1 PYTHON SCRIPT FOR EACH NODE
      When writing or editing the python script for any node's task, follow these guidelines:
      - The script should have the functions preprocess(), compute(), and save(). 
      Those functions may internally call other functions written to make the script well structured for the task.
      Use of smaller functions will also help unit testing.
      - the script has a "state" variable which is initialized to an empty dictionary
      - the preprocess function takes as input a dictionary that maps prior nodes' names to their output or empty dictionary if a prior did not complete. 
      This function should check if the conditions needed to do the task is there; this includes checking for sufficient inputs, and access testing if relevant.
      **IMPORTANT** Always FIRST read the predecessor nodes scripts py files (located in their respective subfolders) to determine what data is in the output of those nodes.
      preprocess function returns true or false, and may update the "state" dictionary if any data preprocessing was done and needed for compute().
      This would be stored under the key "local" inside the "state" dictionary, and the value of "local" should be another dictionary.      
      - the compute function does the task and saves the needed outputs in a dictionary of variable name to value. This is stored in the "state" under the key "output". 
        The output will also have a "task_status" key which is "success" if the task is completed successfully after validation, and "failed" if not. If failed, then an "error_log" key should be in the output with the error message.
      - the save function will pickle the "output" from the state variable for use in the successor nodes. 
      - Include print statements that append to a file named <task_node_name>_summary.txt in the node's subfolder, which should be created at the start of preprocess().
        The first print statement should be the task description that the code does in detail but concise.
        The statements printed to this file should include the main steps completed or errors. It should also include any pertinent results or computed values if it is less
        than 80 characters to print out. if it is a data frame or file created, then just add statements like "file.extension" created.
      - If the script is run , then under "if __name__ == "__main__", it should read the prior nodes' pickle file(s) from their respective subfolders and produce the dictionary of node to output. 
      output is set to empty dictionary if not present in the folder. 
      The folder that contains the prior node's pickle file is: <process_folder>/<prior_node_name>/<prior_node_name>.pkl      
    2.2 UNIT TESTING FOR EACH NODE
      - in the test folder inside the NODE'S SUBFOLDER (make if not there), unit tests for each node should be added. These should be named as test_<node_name>.py
      - unit tests should cover the functionality of the code that can be tested without large data. 
      - any files that need to be created or used for testing should be placed in the test folder

3. FILE NAMING & FOLDER STRUCTURE (CRITICAL)
   - ALL files for each node MUST be placed in the node's subfolder: <process_folder>/<node_name>/
   - Name the main script: <node_subfolder>/<task_node_name>.py (lowercase, non-alphanumeric -> underscore)
      the name will be in the process json under "nodes". The initial key is just the node number, that is not the name. It will be one level deeper under "name" key. 
   - Check the node's subfolder for existing scripts with similar names
   - Reuse/adapt existing code, then save as <node_subfolder>/<task_node_name>.py
   - One Python file per task(node), stored in the node's subfolder
   - Example folder structure:
      process_outputs/
        my_process/
          node_1/                    <- Node 1's subfolder
            node_1.py
            node_1.pkl
            node_1_summary.txt
            test/
              test_node_1.py
          data_analysis/             <- Node "data_analysis"'s subfolder
            data_analysis.py
            data_analysis.pkl
            data_analysis_summary.txt
            output_chart.png         <- Additional outputs go here too
            test/
              test_data_analysis.py

4. TASK DEPENDENCIES
   - Read prior nodes' code to understand their outputs/data. The file is in their subfolder: <process_folder>/<prior_node_name>/<prior_node_name>.py
   - Prior nodes' pickle files are at: <process_folder>/<prior_node_name>/<prior_node_name>.pkl
   - Review any successor nodes' tasks to determine what results they may need for their computation.

5. OUTPUT REQUIREMENTS (CRITICAL)
    Each node's code must:
   - Atleast save results to <node_subfolder>/<task_node_name>.pkl (same name as .py file, in the node's subfolder)
   - Pickle file must contain a dictionary with:
     * "task_status": "success" (after output validation) or "failed"
     * "error_log": (if status is "failed")
     * All pertinent results as key-value pairs
   - Always overwrite previous pickle files with same name
  - If the node's task logic refers to itself (prior computations), the code should read the existing pickle first.
   - Additional output files allowed (in the node's subfolder), but pickle file is mandatory


6. VALIDATION & DEBUGGING
   - Run and debug any Python code written. 
   **IMPORTANT** When running a python script, always tell the script to output stdout and stderr to dedicated files in the project or node subfolder (whichever is more appropriate) to read and delete afterwards.
   - all such files, and temporary code that is not needed for the process and not output files for the process should have the prefix "temp_" in the file name to indicate they are temporary files that should be deleted after reading.

7. IMPORTANT
""" + generate_critical_instructions(working_dirs) + f"""
COMMAND FORMATS:

Remember to issue only one command block at a time, you will be given the result of the command to continue.

Shell commands (SINGLE LINE ONLY - no multi-line commands):
{SHELL_CODE_BLOCK_START}
your_single_line_command_here
{CODE_BLOCK_END}
Use shell syntax appropriate to OS:
- Windows: cmd or PowerShell commands
- Linux/macOS: bash/sh commands
Do NOT use shell commands for file creation, reading, or deletion. Use the dedicated file operation blocks below.

File operations (use these INSTEAD of shell commands for any file I/O):

Read a file:
{FILE_READ_BLOCK_START}
/absolute/path/to/file
{CODE_BLOCK_END}

Write/create a file (first line is the path, everything after is file contents):
{FILE_WRITE_BLOCK_START}
/absolute/path/to/file
file contents go here
line 2 of file contents
...
{CODE_BLOCK_END}

Delete a file:
{FILE_DELETE_BLOCK_START}
/absolute/path/to/file
{CODE_BLOCK_END}

EXECUTION GUIDELINES:
- Use absolute file paths
- Verify work by listing files or viewing content
- Say {TASK_COMPLETED_STRING} when task is complete
- LLM/Embedding code template provided below:

```python
{python_code_for_llm_and_embedding}
#end of python code
```
Use call_llm, get_embedding, or process_image functions as needed. Modify prompts/arguments only.


""" + generate_critical_instructions(working_dirs) 
    #this is based on research that repeating prompt at the end helps 



    return base_prompt

#===============================================
def generate_command_security_prompt(command: str, allowed_folders: list) -> str:
    """
    Generate a prompt for the COMMAND_SECURITY_MODEL to evaluate whether a
    shell command is safe to execute.
    The dedicated LLM is asked to reason about the command and return a structured
    verdict: SAFE or UNSAFE with an explanation.
    This is only a partial measure and insufficient by itself. It should always be used with robust isolation and sandboxed code-execution.
    Args:
        command: The shell command to evaluate
        allowed_folders: List of absolute folder paths the agent is allowed to
                         read/write within

    Returns:
        The complete security-evaluation prompt string
    """
    folders_list = "\n".join(f"  - {f}" for f in allowed_folders)

    return f"""You are a command-security reviewer.  Your ONLY job is to decide
whether the following shell command is **safe** to execute given the security
policy below.  Think step-by-step, then give your final verdict.

=== SECURITY POLICY ===
1. The command MUST NOT perform destructive operations on the host system
   (e.g. formatting disks, fork bombs, shutting down/rebooting, deleting root
   directories, recursive force-delete outside allowed folders).
2. The command MUST NOT read from or write to any path **outside** the allowed
   folders listed below (including via redirections >, >>, <, piping to files,
   or environment-variable expansion that could escape the folder).
3. The command MUST NOT exfiltrate data (e.g. curl/wget uploading files to
   external servers, sending data over the network, or encoding secrets).
4. The command MUST NOT escalate privileges (e.g. sudo for dangerous ops,
   chmod 777 on system dirs, modifying /etc, Windows registry edits).
5. The command MUST NOT install system-wide packages or services that persist
   beyond the current task (apt/yum/brew install of system daemons, Windows
   service registration, etc.).  Installing Python/Node packages inside the
   allowed folders or a virtual environment is acceptable.
6. The command MUST NOT modify or delete files belonging to other processes or
   nodes outside the allowed folders.

Allowed folders (absolute paths):
{folders_list}

=== COMMAND TO EVALUATE ===
{command}

=== INSTRUCTIONS ===
1. Identify every file path (explicit or implicit) the command touches.
2. Determine if any path resolves outside the allowed folders.
3. Check for any dangerous patterns (rm -rf /, fork bombs, disk formatting,
   shutdown, privilege escalation, data exfiltration).
4. Provide your reasoning in 2-4 sentences.
5. End with EXACTLY one of these two lines (no extra text on the line):
   VERDICT: SAFE
   VERDICT: UNSAFE - <short reason>

Your response:
"""
#===============================================
def generate_repeated_core_instructions_for_suffix(working_dirs: list) -> str:
   """
      Repeating some core instructions at the end helps the agent better adhere to instructions
   """
   suffix_prompt = f"""Remember the following
   - Provide regular messages to the user about what you are doing and what you have done, 
     even if they don't ask for it. This is important for keeping the user in the loop and building trust. Use the MESSAGE TO USER format described above.
   - Do NOT read inside the folder .chat_agent_logs         
   - If you create any temporary files for debugging or logging, have the string "temp_" in it and make sure to delete it after reading   
   - If you generate html files for user display/input purposes, try to have the html files be self-contained (include css and js in the same file).
   - Do not directly read binary data as text (including pickle files). Use Python scripts to inspect pickle/binary/large files and extract relevant parts.
   - If you have to inspect a large file, use code to inspect and extract pertinent parts or sections.
  - Use {FILE_WRITE_BLOCK_START} blocks to create or write files. Use {FILE_READ_BLOCK_START} blocks to read files. Use {FILE_DELETE_BLOCK_START} blocks to delete files. Do NOT use shell commands (echo, cat, type, etc.) for file I/O.
  - Shell command blocks ({SHELL_CODE_BLOCK_START}) must contain ONLY A SINGLE LINE command. No multi-line commands, heredocs, or inline file creation via shell.
   - Make sure each command is in the correct block type that starts with {SHELL_CODE_BLOCK_START} (for shell), {FILE_READ_BLOCK_START} (for reading), {FILE_WRITE_BLOCK_START} (for writing), or {FILE_DELETE_BLOCK_START} (for deleting) and end with {CODE_BLOCK_END}, so that they can be parsed and executed by the system.
   - Only ONE command block per response. Do not combine multiple blocks.
   - **ENSURE YOU RUN THE PYTHON CODE AFTER IT IS WRITTEN TO THE SCRIPT FILE TO DO THE TASK, DO NOT OUTPUT PRE-COMPUTED RESULTS**. 
   - **THE PYTHON CODE MUST GENERATE THE .pkl FILE WITH THE OUTPUTS. Verify the code behaved as expected**.
   - Do NOT ask the user to paste code or run commands, do it directly with the appropriate command blocks, or say that you could not do a task and why in plain english in your message to the user.
   - Once your task is completed remove any unnecessary files that were not explicitly asked for or needed in the task. 
   """
   return suffix_prompt
\end{lstlisting}


\end{document}